# Robotic needle steering in deformable tissues with extreme learning machines

P. H. Suruagy Perrusi[1*], A. Cazzaniga[2], P. Baksic[1], E. Tagliabue[3], E. De Momi[2], H. Courtecuisse[1]

[1] *Laboratory ICube Équipe Automatique Vision et Robotique (AVR), Strasbourg, France*
[2] *Electronic Information and Bioengineering Department, Politecnico Di Milano, Milan, Italy*
[3] *Altair Robotics Lab, Computer Science Departement, University of Vernona, Verone, Italy*
[*] *Corresponding author, email: phsuruagyperrusi@unistra.fr*

*Abstract:* Control strategies for robotic needle steering in soft tissues must account for complex interactions between the needle and the tissue to achieve accurate needle tip positioning. Recent findings show faster robotic command rate can improve the control stability in realistic scenarios. This study proposes the use of Extreme Learning Machines to provide fast commands for robotic needle steering. A synthetic dataset based on the inverse finite element simulation control framework is used to train the model. Results show the model is capable to infer commands 66% faster than the inverse simulation and reaches acceptable precision even on previously unseen trajectories.

## I. Introduction

Robotic assistance for needle-based therapies has been the subjects of many study[1], [2]. Given the high nonlinearity of the needle and tissue interaction, the precise control of these robots remains an open challenge. A promising control framework which accounts for deformations of both the needle and the tissue is the inverse finite element (FE) simulation [1], [2]. It is an automatic needle insertion strategy to steer needles at a constant velocity along a pre-defined trajectory inside soft tissues. The needle-tissue interaction is compensated via optimization in a realistic FE simulation. In its latest formulation [2], reduced computational time increased the dynamic of the robotic allowing for more stability when subject to perturbations.

On another hand, recent Machine Learning models can learn to predict non-linear soft tissue deformations with very low computation time [3]. Since patient-specific organ deformation data is uncommon, they rely on FE simulations to generate their training set. The synthetic data demonstrated high representativity of the physical scenario as they were able to estimate accurate elastic deformations 500 times faster than using FE simulations.

In particular, the Extreme Learning Machine (ELM) [4] is a neural network architecture which presents promising results in solving the inverse kinematics of robots with high non-linearities between the control signal and their position [5]. We propose here to solve an analogous problem, that is: solving the inverse kinematics linking the needle tip motion to the robot end-effector motion using ELM, while considering the needle-tissue interaction.

## II. Material and methods

In this study we propose to predict robotic position commands which account for needle and tissue deformations by using ELM. The proposed network is trained with synthetic data generated from inverse finite element (FE) simulations of robotic needle insertions within the SOFA Framework [6]. Finally, the model's inference is used to drive a robotic needle insertion in a FE simulation with increased command dynamics.

### II.I. Inverse Simulation

In [2], the authors formulate the needle insertion as a minimization problem. A realistic FE simulation is used to linearize the relationship between the end-effector pose and a set of objective functions. This linearization is then used to derive a robot pose command. A first objective $e = P_{target} - P$ minimizes the error between the needle tip position $P$ and a target position that moves along a pre-defined trajectory $P_{\text{target}}$. The other two objectives reduce tissue damage at the level of the entry-point and enforce a remote center of motion (RCM) constraint to the optimization problem.

### II.II. Dataset generation

The training dataset is generated using realistic FE simulations of robotic needle steering in a rectangular-shaped foam, as in [1]. FE models use linear co-rotational formulation of elasticity. The needle is modeled as serially linked beam-elements following Timoshenko's formulation. The needle tissue FE interaction follows [7].

In order to explore multiple needle-tissue interaction scenarios, a set of 105 straight trajectories were created. Each trajectory passes through the insertion point at the center of the *XY* plane of the foam. Their endpoints are distributed in a regular grid at insertion depth of $110 mm$.

For each trajectory, the inverse simulation control of needle insertion is performed. In order to facilitate the generalization to new unseen trajectories, isotropic uniform noise of $\pm 0.25mm$ is added to $P_{target}$ during the data generation. Finally, for each sample $i$, the current effector 3D position $E_{ff_{(i)}}$, the current needle tip to target error $e_{(i)}$ and respective robotic displacement output $C_{(i)}$ are saved. The complete dataset generated from the inverse simulations on the straight trajectories is composed of 100000 samples, in which 90% is used as training set and 10% are left for as test set.

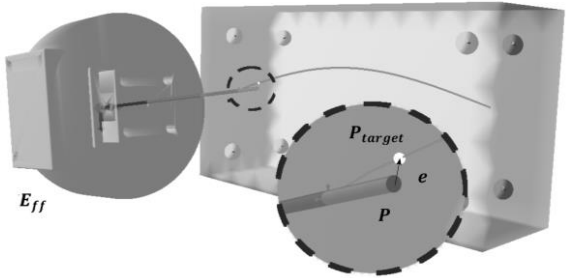

*Figure 1: FE simulation with a curved trajectory used to validate the ELM needle steering. The robotic end-effector position $E_{ff}$ and the needle tip-to-target error $e = P_{target} - P$ are the model inputs.*

### II.III. Neural Network Architecture

ELM is a feed forward neural network with a single hidden layer. In this study it was implemented with 25 neurons with sigmoid activation functions. This model maps the inputs $x_i = \begin{bmatrix} E_{ff_{(i)}} & e_{(i)} \end{bmatrix}^T$ into displacements of the robotic end-effector: $y_i = [C_{(i)}]$.

It is assumed the $x_{(i)}$ can be measured at each time step. In this study, the ELM will be tested in a realistic FE simulation in the same scenario used to generate the dataset. The robotic end effector translates at a constant speed towards the inferred position $C$, while its orientation is imposed to respect an RCM between the needle base and the entry-point.

## III. Results and discussion

In a first phase, the ELM is trained and validated with synthetic data acquired from inverse simulations (see sec. II.II). Results on the synthetic data show a test phase final root mean squared error (RMSE) of $0.164\ mm$. The trained weights of the ELM are then conserved for the remaining experiments.

In a second phase, its ability to steer a needle in a FE simulation is tested. A computation time analysis shows the ELM inference takes $0.91 \pm 0.67ms$ to perform predictions while the inverse simulation needs $2.69 \pm 0.34ms$. When comparing the tip-to-target precision in all trajectories, the ELM follows the straight trajectories with a final RMSE of $91 \pm 23\mu m$. while the inverse simulation presents $70 \pm 1\mu m$. It remains to be tested which precision the method achieves in a realistic scenario such as [1], [2].

In order to test the ELM generalization capability on previously unseen trajectories, an additional set of 200 curved trajectories were generated. They present a $10mm$ sinus overlay to the straight trajectories in X and Y directions. As a result, the ELM remained stable and performed the needle insertion with final RMSE of $213 \pm 517\mu m$. The inverse simulation followed these same trajectories with a final RMSE of $818 \pm 108\mu m$. Higher errors on both methods were expected as more significant deformations are imposed to follow a curved trajectory [1].

As a first feasibility study, results show that the ELM was able to learn how to perform stable needle insertions, achieving a reduction in the average computational time by 66% when compared to the inverse simulation. This performance boost might improve the dynamics of the control system in a real application. The precision observed in the experiments is sub-millimetric and remains acceptable for percutaneous needle insertion procedures. In terms of limitations, this study assumes constant biomechanical parameters and simplistic geometry for the models both in the synthetic data generation and in the experiments.

## IV. Conclusions

This study presents a first feasibility study to train a neural network to control robotic needle insertions in soft tissues and achieve appropriate precision in a realistic simulation environment. The ELM presented a promising ability to account for tissue deformation and trajectory generalization ability for the straight and curved trajectories tested. Future works will build upon these findings to improve insertion precision and transfer to different scenarios. A sensitivity study of network performances when varying such parameters will be carried out in future works. An experimental validation in a controlled environment will also be an important milestone to advance the studies towards a clinical environment.

**AUTHOR'S STATEMENT**

Research funding: This work was supported by French National Research Agency (ANR) within the project SPERRY ANR-18-CE33-0007 and the Investissements d'Avenir program (ANR-11-LABX-0004, Labex CAMI). Conflict of interest: Authors state no conflict of interest.